# Induction and Uncertainty Management Techniques Applied to Veterinary Medical Diagnosis

*by*

**M. Cecile**
**M. McLeish**†

*Department of Computing and Information Science*
*University of Guelph*
*Guelph, Ontario, Canada   N1G 2W1*

**P. Pascoe**          **W. Taylor**
*OVC, Guelph*    *NASA AMES, Moffett Field,*
*CA 94035*

## Abstract

This paper discusses a project undertaken between the Departments of Computing Science, Statistics, and the College of Veterinary Medicine to design a medical diagnostic system. On-line medical data has been collected in the hospital database system for several years. A number of induction methods are being used to extract knowledge from the data in an attempt to improve upon simple diagnostic charts used by the clinicians. They also enhance the results of classical statistical methods - finding many more significant variables. The second part of the paper describes an essentially Bayesian method of evidence combination using fuzzy events at an initial step. Results are presented and comparisons are made with other methods.

## 1. Introduction

This paper discusses the progress to date on a large project undertaken primarily at the University of Guelph, which is the home to one of Canada's major (and few) Veterinary Colleges. The larger goal of the project is to build a general diagnostic shell for veterinary medicine. Work has begun on a prototype involving the diagnosis of surgical versus medical colic in horses. This is a significant problem in veterinary medicine having led to many studies, use of diagnostic charts etc. to aid owners and veterinarians in recognizing serious cases[6,7,24]. Horses suspected of requiring surgery must be shipped at a significant cost to the veterinary hospital, where further tests are conducted and a final decision is made. We are endeavoring to provide a computerized diagnostic tool for use in the hospital as well as on a remote access basis for practicing veterinarians.

The animal hospital at Guelph has a computer system call VMIMS (Veterinary Medical Information Management System). This system was originally programmed in Sharp APL and handles usual admission and billing procedures. However, the system also stores a considerable amount of medical information on each patient, including bacteriology, clinical pathology, parasitology, radiology and other patient information such as age, sex, breed, presenting complaint, treatment procedures, diagnosis and outcome. In clinical pathology, much of the data is electronically generated by the lab equipment. The database currently holds about 48,000 records with 30,000 unique cases, requiring 500 megabytes of disk storage. The current project is being designed to make use of the vast amounts of on-line data to aid the diagnostic process. Other database issues concerning the implementation of our system are discussed by M. McLeish, M. Cecile, and A. Lopez in [12]. A diagram of the proposed system is given in [13].

† This research has been supported by NSERC operating grant #A4515.

Medical expert systems have been under development for human medicine for many years, especially as seen in major projects like MYCIN (Stanford) [3]. These recent systems have been largely based on the assumption that to have expert capability, they must somehow mimic the behavior of experts. Earlier work, using mathematical formalisms (decision analysis, pattern matching, etc.) were largely discarded and attention to the study of the actual problem-solving behavior of experienced clinicians. In a recent paper [21] by Drs. Patil, Szolovits and Schwartz, it is suggested that the time has come to link the old with the new: "now that much of the A.I. community has turned to casual, pathophysiologic reasoning, it has become apparent that some of the earlier, discarded strategies may have important value in enhancing the performance of new programs ...." The authors recognize the difficulty of this approach when they state that "an extensive research effort is required before all these techniques can be incorporated into a single program."

This project is truly an attempt to allow rules and other types of information extracted from data to enchance or complement an "expert opinion" driven system. The paper discusses our current progress with induction processes and uncertainty management methods.

Classical statistical methods of discriminant analysis and logistic regression were run on data sets involving about 45 input parameters. The results of these analyses are summarized in section 2.1. Problems with the predictive power and variable selection of these methods are discussed. Reasons are provided for looking at other methodologies to help deal with these problems.

Section 2.2 discusses the use of Bayesian classification and how the results of this method can extend the information discovered in section 2.1. It also outlines how Bayesian classification may be used predictively for the type of problem at hand. Section 2.3 briefly describes some other induction methods which have been applied to the data.

Section 3 outlines a method of evidence combination which is based on a form of Bayesian updating but uses infinite valued logic as a basis for the representation of events. Some initial results of its use are tabulated. Finally section 4 highlights work which is planned and currently in progress.

### 2.1 Classical Statistical Methods

The data used for these studies represented 253 horses presented at the teaching hospital at Guelph. The horses were all subjected to the same clinical tests and the same pathology data were collected. This data set was used for all the studies discussed in this paper. Outcome information of the following type was available: whether surgery was performed, whether or not a surgical lesion was actually found and the final state of the animal (i.e. lived, died, or euthanized). The objective of the study was to assess which variables obtained during examination of the horses with abdominal pain were significant in differentiating between horses that required surgery versus non-surgical treatment. The types of parameters which were studied included rectal temperature, heart rate, respiratory rate, temperature of extremities, colour of mucous membranes, capillary refill times, presence and severity of abdominal pain, abdominal distension, peristalsis, and the results of naso-gastric intubation and rectal examination. Clinico-pathological parameters evaluated included hematocrit (HCT) and the total plasma concentration of abdominal fluid. These variables were sometimes continuous and when descriptive (pain levels, etc.) were translated into discrete integer variables. Missing data was handled by elimination of cases. There are 20 parameters in each of the two data sets (*clin* and *clin-path*).

A multiple stepwise discriminant analysis in a recursive partition model was used to determine a decision protocol. The decision protocol was validated by a jackknife classification and also by evaluation with referral population in which the prevalence of surgical patients was 61% [c.f. 6, 7]. The significant parameters were found to be abdominal pain, distension and to a lesser extent, the color of abdominal fluid. The use of the decision tree yielded a significant number of false positives and virtually eliminated false negatives in one study. Unnecessary surgery is even more undesirable in animals than humans due to costs (usually borne by the owner) and the debilitating effects of surgery on a productive animal. Other difficulties with these results concerned the fact that the clinical pathology data appeared entirely non-predictive - a result contrary to the medical belief that, at least in serious cases, certain of these measured parameters do change significantly. Discriminant analysis can miss effects when variables are not linearly behaved. Missing data was another serious problem. Other methods described in section 3 helped overcome some of these problems.

Logistic Regression [11] was also run on the same data set. Here, a regression model of the form
$$Y_i = \log_e [\frac{P_i}{1-P_i}] = \beta_0 + \beta_1 X_1 + \cdots \beta_k X_k$$
is used where the $\beta_i$'s are slope parameters relating each of the $X_i$ independent variables to the $Y_i$'s (log odd's ratio). $P_i$ is the probability of a response and can be estimated by the back transformation: $P = \frac{e^Y}{(1-e^Y)}$. When a new case is to be diagnosed, the probability of requiring surgery may be calculated. The data was run for all three possibilities: surgical lesion found (SL), surgery performed (S) and outcome (O). The outcome, O, can take on the values of lived, died or euthanized. Pulse, distension and a variable representing the presence of firm feces in the large intestine (A2) were the significant predictors. However, testing against whether the doctors decided to do surgery, pain and A2 were the most significant. (These results were obtained from the clinical data only.) Outcome found several other variables to be significant; the probability of death is increased by pain, cold extremities, a high packed cell volume and low NGG reading (Naso-gastric tube emissions). Again, problems were caused by the presence of missing data. Some modifications of these results were found when missing data were estimated using stepwise regression methods. The reasons for the missing data were complicated and various - sometimes records ended because euthanasia was chosen as a solution for cost reasons. Thus, estimating missing data was not a very reliable technique.

### 2.2 Bayesian Classification

This methodology uses Baye's Theorem to discover an optimal set of classes for a given set of examples. These classes can be used to make predictions or give insight into patterns that occur in a particular domain. Unlike many clustering techniques, there is no need to specify a "similarity" or "distance" measure or the number of classes in advance. The approach here finds the most probable classification given the data. It allows for both category-valued information and real-valued information. For further details on the theory, the reader is referred to [4,5,22].

A program called Autoclass I (see also [5]) was run on the combined clinical and pathological data sets. All 51 variables were included; that is, all outcome possibilities, as described in section 2, were included as variables and lesion type (four possibilities) was also added. A total of 13 classes were found and in most cases the probabilities of a horse belonging to a class were 1.00. The type of information available consisted of relative influence values for every attribute to the over-all classification. For each class and each attribute, influence values are produced indicating the relative influence of each attribute to that class. This information is available in tabular and graphical form.

The classes provide related groups of cases which are useful for case studies (OVC is also a teaching institution). The information may be used predictively. For example, the class with the highest normalized weight, class 0, was found to have surgical lesion as a very high influence factor. The variables of abdominal distension, pulse, abdomen (containing A2 mentioned earlier) and pain, found significant by earlier methods, were also influential factors for this class. Some other variables not flagged by earlier methods were found to be influential as well (total protein levels and

40

abdominocentesis, in particular). The horses in class 0 were found to not have surgical lesions. It is thus possible to see from the features of horses in this class, which attributes and what type of attribute values are significant for this to be the case. New cases can be categorized into classes according to their attribute values and if found to be in class 0, this would indicate a very small chance of surgery being required. Class 1, however, is predominantly a class where surgeries are required.

Actually, there is a wealth of information to be gleaned from the results and much of this interpretive work is ongoing at this time. One may infer that certain variables are not very predictive. For example, a variable distinguishing young from old animals has low influence value in all major classes, but is slightly more significant in a couple of the smallest classes. Studying these small classes, however, can be particularly fascinating because they flag situations which are more unusual. That is, methods which simply find variables which are *usually* the most predictive, cannot perform well on cases which do not conform to the normal pattern. For example, class 11 has four cases all of which required surgery. A number of pathology variables were the most influential, with pain, abdomen, and distension being only moderately important. Surgical type was a very influential variable and indeed three of the cases actually had a strangulated hernia involving the small intestine and one a strangulation problem of the small intestine called intussuception. (The number of possible precise diagnoses was approximately 12x4x3x11.) This class was also interesting because two of its cases were young (less than nine months). The percentage of young cases in the whole population was less than 8%.

Other classes pinpoint cases difficult to diagnose. Class 12 contains three cases which were all operated on, but only one out of three was actually found to have a surgical lesion. Two cases had a simple large colon obturation and the third a large colon volvulus or torsion (requiring surgery). That is, two unnecessary surgeries were performed. A close study of the influential variables and their parameter values for this class of cases - with very close but importantly different diagnoses - provides extremely valuable information.

A significant question arises as to whether or not outcome information should be included initially in the program run. If the data were highly predictive, it should not matter. However, it was not clear from the onset how true this was. The data was run twice again: once with all outcome information removed and once with the doctor's decision information deleted. When all outcome information was removed, one notes that some interesting prognosis information still remains. Class 1 indicated cases virtually all of which lived and whose condition, whether or not surgery was required, was generally good. The question often arises whether or not to operate even given that the animal has a lesion if the general prognosis is bad. This class would indicate that surgery should be performed in such cases. Class 2 (42 cases) was extremely well discriminated with 91% having a surgical lesion. Of the remaining classes (again there were 13 in total), 3/4 were also reasonably well discriminated on the basis of lesion. Others flagged items such as young animals or cases that had very poor over-all prognosis.

New data has been obtained and code written to take new cases and determine a probability distribution, for this case over the classes, from which a probability of outcome may be calculated. It is interesting to note that the use of Bayesian classification for medical diagnosis in this fashion is in a sense a mathematical model of the mental process the clinicians themselves use. That is, they try to think of similar cases and what happened to those cases in making predictions. The technique for making predictions outlined above provides a sophisticated automation of this process. It is impossible within the scope of this paper to document all the information obtained from using Bayesian inductive inference. A sample case will be provided in Section 3.5 which will show the usefulness of combining information from the results of Bayesian inductive inference with the other methodologies.

## 2.3 Other Induction Techniques

Several other methods have been tried on the data sets. The probabilistic learning system [c.f. 19, 20] also classifies data. Classes are optimally discriminated according to an inductive criterion which is essentially information-theoretic. To accommodate dynamic and uncertain learning, PLS1 represents concepts both as prototypes and as hyperrectangles. To accommodate uncertainty and noise, the inductive criterion incorporates both probability and its error. To facilitate dynamic learning, classes of probability are clustered, updated and refined. The references describe the methodology in detail. We give here some of the results of the application to the veterinary data.

Two classes of data were given to the program (**1**: no surgery, **2**: surgery) and 14 clinical pathology variables were processed according to the PLS1 algorithm. The results can be interpreted as rules and also flag the most prominent variables. (The data sets were reduced due to the presence of missing data. Dr. Rendell is currently revising the PLS1 algorithm to allow it to run with a minimum loss of data when missing values are present.) The current version of PLS1 required that the data be scaled between 0 and 255. The variable X1 represents total cell numbers, X8 is mesothelial cells and X13 is inflammation. These were found to be the most significant variables for the prediction of no surgery. For the purpose of predicting surgery, again X1 and X13 were significant, as well as X9, a measure of degenerate cells. Uncertain rules can also be obtained from the results. Further work is being done to revise the learning algorithm to be able to handle missing data without losing entire cases (i.e. all parameter values when only one is absent). The statistical methods lose all case information and it would be a considerable advantage to employ a method less sensitive to this problem.

Quinlan's algorithm [18] for rule extraction has been run on the clinical data. Continuous data has been converted to discrete values. The important variables are mucous membranes, peristalsis, rectal temperature, packed cell volume, pulse, abdominal condition, and nasogastric reflux. A number of variables appeared significant using this technique which were deemed unimportant using discriminant analysis. A decision tree was generated but some implementational difficulties, due to the number of variables used and missing data, led to a tree which was not very complete or reliable. These problems are currently being addressed to develop a more robust version of the algorithm.

Further details and output on the methods just discussed can be found in [14]. The method in the following section uses all variables provided, rather than reducing the variable set and basing the diagnoses on only a few parameters.

## 3.1 Overview

This section describes a method of evidence combination performs Bayesian updating using evidence that may be best modelled using an infinite valued logic such as that which fuzzy set theory provides. The methodology described in this section provides a unified approach for intelligent reasoning in domains that include probabilistic uncertainty as well as interpretive or "fuzzy" uncertainty.

A formulation central to several components of the methodology is that of the "weight of evidence" and is therefore introduced in section 3.2. The description (and justification) of the use of an infinite valued logic is presented in section 3.3. Section 3.4 explains how "important" symptom sets are discovered and section 3.5 relates the performance of this method in the domain of equine colic diagnoses.

## 3.2 The Weight of Evidence

A.M. Turing originally developed a formulation for what he called the "weight of evidence provided by the evidence E towards the hypothesis H" or W(H:E). Good [8,9] has subsequently investigated many of the properties and uses of Turing's formulation which is expressed as:

$$W(H:E) = \log\left(\frac{p(E/H)}{p(E/\overline{H})}\right) \quad Or \quad W(H:E) = \log\left(\frac{O(H/E)}{O(H)}\right)$$

42

where O(H) represents the odds of H, $\frac{p(H)}{p(\overline{H})}$. Weight of evidence plays the following part in Bayesian inference:

$$\text{Prior log odds} + \sum_i weight\ of\ evidence_i = \text{posterior log odds}$$

A weight of evidence which is highly negative implies that there is significant reason to believe in $\overline{H}$ while a positive W(H:E) supports H. This formulation has been most notably used in a decision support system called GLADYS developed by Spiegelhalter [23].

In any formulation for evidence combination using higher order joint probabilities there exists the problem of evidence that may appear in many different ways. For example, a patient has the following important symptom groups: (High pain), (High pain, high temp.), (High pain, high temp, high pulse). Which of these symptom groups should be used? Using more than one would obviously be counting the evidence a number of times. The rule we have chosen to resolve this situation is to choose the symptom group based on a combination of the group's size, weight, and error. In this way we may balance these factors depending upon their importance in the domain. For example, if higher order dependency is not evident in a domain then the size of a group is of little importance.

### 3.3 Events as Strong $\alpha$ - Level Subsets

Infinite valued logic (IVL) is based on the belief that logical propositions are not necessarily just true or false but may fall anywhere in [0,1]. Fuzzy set theory is one common IVL which provides a means of representing the truth of a subjective or interpretive statement. For example, what a physician considers to be a "normal" temperature may be unsure or "fuzzy" for certain values. For these values the physician may say that the temperature is "sort of normal" or "sort of normal but also sort of high". This is different from the likelihood interpretation where the probability of a normal temperature in certain ranges is between 0 and 1. Implicitly, probability theory (in both the belief and frequency interpretations) assume that an event either happens or does not (is true or false). On the practical side, we have found that the concept and estimation of membership functions is intuitively easy for physicians.

Let F be a **fuzzy subset** of a universe, U. F is a set of pairs $\{x, \mu_F(x), x \in U\}$ where $\mu_F(x)$ takes a value in [0,1]. This value is called the grade of membership of x in F and is a measure of the level of truth of the statement "x is a member of the set F".

A strong $\alpha$ - **level subset**, $A_\alpha$, of F is a fuzzy set whose elements must have a grade of membership of $>= \alpha$. Formally defined,

$$A_\alpha = \{\mu_{A_\alpha}(x) \mid \mu_{A_\alpha}(x) >= \alpha\}$$

For example, if we have the fuzzy set $F = \{x1/0.2, x2/0.7, x3/0.0, x4/0.4\}$ then the strong $\alpha$ - level set $A_{\alpha=0.2} = \{x1/0.2\ x2/0.7, x4/0.4\}$.

Because we wish to perform probabilistic inference we need to have a means of calculating the probability of fuzzy events. Two methods have been suggested for this: the first from Zadeh [26] and the second from Yager [25]. Zadeh's formulation is as follows:

$$P(A) = \int_{R^N} U_A(x) dp = E[U_A(x)]$$

$U_A$ is the membership function of the fuzzy set A, and $U_A \in [0,1]$. Yager argues that "... it appears unnatural for the probability of a fuzzy subset to be a number.". We would further argue that Zadeh's formulation does not truly provide a probability of a fuzzy event but something quite different: the expected truth value of a fuzzy event. Yager proposes that the probability of a fuzzy

43

event be a fuzzy subset (fuzzy probability):

$$P(A) = \bigcup_{\alpha=0}^{1} \alpha \left[ \frac{1}{P(A_\alpha)} \right]$$

where $\alpha$ specifies the $\alpha$ - level subset of A and since $P(A_\alpha) \in [0,1]$, P(A) is a fuzzy subset of $[0,1]$. This fuzzy subset then provides a probability of A for every $\alpha$ - level subset of A. Thus, depending on the required (or desired) degree of satisfaction, a probability of the fuzzy event A is available. In our case the desired level of truth is that which maximizes the the bias of this event to the hypothesis. For example, if we wish to set a degree of satisfaction for the proposition "x is tall" and we are primarily interested in whether x is a basketball player then we wish to choose an $\alpha$ level which allows us to best differentiate BB players from non-BB players. We define this optimal $\alpha$ - level to be:

$$\underset{\alpha}{Max} \left| W(H{:}E_\alpha) \right| \quad \alpha \in [0,1]$$

$W(H{:}E_\alpha)$ is the weight of evidence of the strong $\alpha$ - level subset $E_\alpha$ provided towards the hypothesis H. The $\alpha$ -level which maximizes the bias of a fuzzy event to a hypothesis (or null hypothesis) is the optimal $\alpha$ - level for minimizing systematic noise in the event.

### 3.4 Discovery of Important Attribute Sets

The identification of important sets of symptoms or characteristics is done commonly by human medical experts and other professionals. For example, the combination of (abdominal pain, vomiting, fever) may indicate appendicitis with a certain probability or level of confidence. Our motivation for trying to discover important symptom groups is twofold: to identify which groups are important in a predictive sense and to quantify how important a group is. Also a factor in the decision of using symptom groups instead of individual variables is the belief that there exists many high order dependencies in this and other real-life domains. For example, a high pulse rate, a high respiratory rate, and high temperature are obviously dependent in many ways that are violations of the Bayesian independence assumption if taken as single symptoms. In using higher order joint distributions for several groups of symptoms we can account for higher order dependencies in the domain and hopefully not violate the assumptions of the Bayesian model.

An symptom set may be of any size between 1 and N where N is number of possible symptoms. To find all such symptom sets requires an exhaustive search of high combinatorial complexity. This may be reduced somewhat by not examining groups that contain a subset of symptoms which are very rare. For example, if freq(High pain, low pulse) is very low then we need not look at any groups containing these two symptoms. Our present implementation examines sets up to size three. The weight of evidence of each symptom group is measured from the data and a test of significance decides whether this group has a weight significantly different from 0.

Of significant interest is the clinician's endorsement of the important symptom sets that this method found. Those sets which showed as being important using the weight of evidence are symptom groups that the clinician would also deem as being significant.

### 3.5 Implementation and Results

Implementation was on a Sequent parallel processor with 4 Intel 80386's. The method was coded in C and Pascal and made much use of a programming interface to the ORACLE RDBMS. This provided a powerful blend of procedural and non-procedural languages in a parallel programming environment.

Data was obtained for a training set of 253 equine colic cases each composed of 20 clinical variables. Also included for each case are several pertinent diagnostic codes: clinician's decision, presence of a surgical lesion, and lesion type. The prototype system provides a prediction for the presence of surgical lesions. Veterinary experts commonly have problems in differentiating between surgical and non-surgical lesions. Of primary concern to the clinicians is the negative predictive value: that is, how often a surgical lesion is properly diagnosed. If a surgical lesion is present and is incorrectly diagnosed then the lesion is usually fatal for the horse. Presented below is a summary of our results using 89 cases from the training set.

| Comparison of Predictive Power (89 Cases) | | |
| --- | --- | --- |
| Method | Negative Predictive Value | Positive Predictive Value |
| Clinician† | 87.6% | 100% |
| Weight of Evidence | 96.7% | 86.6% |

† Figures obtained from these 89 cases. Previous studies have shown these values to be 73% and 93% respectively over a large sample.

From these results we can see that the method of evidence combination achieved an accuracy for negative prediction which exceeded the clinician's. Incorrect diagnoses are being reviewed by the clinician to see if some explanation can be found. There seems to be no correlation between clinician's errors and the computer technique's. This perhaps indicates that the clinician is adept at cases which are difficult for our techniques (and vice versa).

The following example shows our results for a case which had a surgical lesion but was not operated on by the clinicians. The horse displayed the following symptoms:

| | | | |
| --- | --- | --- | --- |
| > 6 months old | High rect temp | Very high pulse | high resp rate |
| Cool temp at xtrem | Reduced per pulse | Norm mucous membranes | Cap refill < 3s |
| Depressed | Hypomotile | Mod abdominal distension | Sl nasogastric reflux |
| No reflux | Elevated Reflux PH | Normal rectal Xam | Distended lg intestine |
| Norm packed cell vol | Normal total protein | serosang abdominocentesis | High abd Tot Protein |

and the method determined that the following evidence was important:

| Evidence in Favor of Surgical Lesion: | |
| --- | --- |
| Symptom Group | W(H:E) |
| > 6 months,Hypomotile,Mod abdom dist | 1.053 |
| Sl nasogastric,Distended L.I.,Normal Tot. Protein | 0.861 |
| V high pulse,Reduced per pulse, C refill < 3s | 0.861 |
| Cool temp Xtrem,No reflux,Normal PCV | 0.661 |
| High rect temp, depressed | 0.372 |
| Serosang abdominocentesis | 0.312 |
| Evidence Against Lesion: | |
| High resp rate,norm mucous mem,normal rectal Xam | -0.547 |

Final Results:

| Prior Log Odds | = 0.530 |
|---|---|
| + W(H:E) | = 3.573 |
| = Post. Log Odds | = 4.103 |

==> p(surg lesion) = 0.804

For this case, this method strongly supports the surgical lesion hypothesis. It is interesting to compare these results to that of the classical regression model. Using this model the probability of a lesion, p, is predicted by: $p = \frac{e^Y}{(1-e^Y)}$, where

$$Y = 7.86 - 1.73(A\,2) - 1.54(ln\,(pulse)) - 0.498(Distension)$$

In this case A2 was 1 because the horse had a distended large intestine, the pulse was 114, and the distension was 3 (moderate). Substituting into the formula we get:

$$Y = 7.86 - 1.73 - 1.54(ln(114)) - 0.498(3)$$
and when we solve for p we find that p = 0.5818

Thus the classical regression analysis produces a p value rather than .5, but which is not strongly convulsive. We speculate that this may be due to the fact that so few variables are considered by this method.

It is interesting to combine these findings with the results of Bayesian classification. This case belongs to autoclass 4 (determined with outcome information excluded). In this class, 80.1% of the 22 cases had a lesion, close to the value predicted but the weight of evidence formula. However, in this class only 19% of the cases lived and only 15% of the animals which had a lesion and were operated on actually lived. Pathology variables were particularly important for determining that class and abdominal distension was only moderately significant (although the doctors and logistic regression rely on this variable). The Autoclass information suggests a poor outcome prognosis in any case and indeed this was a situation in which the clinicians decided on euthanasia. Regression and weight of evidence techniques alone would not have suggested this decision.

## 4.0 Conclusions and Further Work

This paper has attempted to provide an idea of the methods being used to extract information from data in the development of an information system for medical diagnoses. Several techniques have been presented and some initial comparisons made.

Some tests of performance were accomplished by keeping aside portions of the test data. A new data set is currently being gathered with 168 cases which will be used to both test the methodologies and then refine the present diagnostic results. This new data set has been difficult to retrieve as not all the data used in the original set was on-line. We are taking care to ensure the information taken from hard copy records is entirely consistent with the first training set.

We are also looking at a more standard Bayesian model and trying to understand the dependencies and other conditioning in the data. The methodologies used here also help shed some light on this. The works by J. Pearl [16] and Spiegelhalter [10] are being considered for this approach.

In terms of developing an actual system using the methodologies of part 3, a prototype has already been implemented which considers symptom groups up to a size of three. A more advanced algorithm which tests independence between groups and provides an error estimate is currently being implemented in a blackboard architecture.

Terminals exist in the hospital in the work areas used by the clinicians and we are now proceeding to make the results available on incoming cases. Any final system would provide the doctors with selected information from several methodologies. This is to help especially with the diagnosis of difficult cases - as the real question is not just to be statistically accurate a certain percentage of the time, but to have diagnostic aids for the harder cases.

### Acknowledgements

The authors wish to thank the Ontario Veterinary College computing group, Dr. Tanya Stirtzinger (OVC) and Ken Howie for their help, especially with data collection. The Statistical consulting group at the University of Waterloo (C. Young) helped with some analyses. The encouragement of Dr. D.K. Chiu is also gratefully acknowledged.